\documentclass[]{elsarticle}
\pdfoutput=1

\usepackage{hyperref}
\usepackage{amssymb}
\usepackage{graphicx}
\usepackage{bbm}
\usepackage{amsmath}
\usepackage{subfig}
\usepackage{algorithm}
\usepackage{algcompatible}
\usepackage{algpseudocode}
\usepackage{booktabs}

\biboptions{authoryear}

\begin{document}

\begin{frontmatter}

\title{Perturb-and-MPM: Quantifying Segmentation Uncertainty in Dense Multi-Label CRFs}


\author[ISTB]{Raphael Meier\corref{cor1}}
\ead{raphael.meier@istb.unibe.ch}
\author[Inselspital]{Urspeter Knecht}
\ead{urspeter.knecht@insel.ch}
\author[ISTB]{Alain Jungo}
\author[Inselspital]{Roland Wiest}
\ead{roland.wiest@insel.ch}
\author[ISTB]{Mauricio Reyes}
\ead{mauricio.reyes@istb.unibe.ch}
\address[ISTB]{Institute for Surgical Technology and Biomechanics, University of Bern, Stauffacherstrasse 78, 3014 Bern, Switzerland}
\address[Inselspital]{Support Center for Advanced Neuroimaging -- Institute for Diagnostic and Interventional Neuroradiology, University Hospital and University of Bern, Freiburgstrasse 4, 3010 Bern, Switzerland}
\cortext[cor1]{Corresponding author}

\begin{abstract}
This paper proposes a novel approach for uncertainty quantification in dense Conditional Random Fields (CRFs). The presented approach, called Perturb-and-MPM, enables efficient, approximate sampling from dense multi-label CRFs via random perturbations. An analytic error analysis was performed which identified the main cause of approximation error as well as showed that the error is bounded. Spatial uncertainty maps can be  derived from the Perturb-and-MPM model, which can be used to visualize uncertainty in image segmentation results. The method is validated on synthetic and clinical Magnetic Resonance Imaging data. The effectiveness of the approach is demonstrated on the challenging problem of segmenting the tumor core in glioblastoma. We found that areas of high uncertainty correspond well to wrongly segmented image regions. Furthermore, we demonstrate the potential use of uncertainty maps to refine imaging biomarkers in the case of extent of resection and residual tumor volume in brain tumor patients.
\end{abstract}


\end{frontmatter}

\newtheorem{prop}{Proposition}
\newproof{pf}{Proof}

\section{Introduction}
Image segmentation plays a pivotal role in the analysis of medical imaging data. Information extracted via segmentation (e.g. volume/position) can be used for diagnosis, treatment planning and monitoring. Graphical models offer a way to describe structure in medical images and embed it in a probabilistic framework. Markov or Conditional Random Fields (CRFs) have become one of the most widely-used graphical models in image understanding \citep{Wang2013}. Although CRFs are probabilistic graphical models (they model a Gibbs distribution), the final image segmentation corresponds usually to the most probable (MAP) labeling (=hard labeling). Sampling from the underlying Gibbs distribution would require the use of Markov Chain Monte Carlo (MCMC) methods, which are computationally very expensive and thus prohibitive for clinical applications. Efficient sampling would enable quantification of \textit{segmentation uncertainty}, which could better support the decision making process of clinicians employing the segmentations. More specifically, image regions that are segmented (automatically) with high uncertainty can be e.g. reassessed by a human expert or excluded from subsequent analysis.

The most popular CRF model corresponds to a structured grid, where voxels are represented by nodes in a graph. For neighboring nodes pairwise potentials are defined which aim to induce smoothness of the image segmentation. The use of a Potts prior in grid-structured pairwise CRFs is known to result in excessive smoothing (also known as \textit{shrinking bias} \citep{Kohli2009,Meier2016b}) of the object boundary. Dense CRFs establish pairwise potentials between all nodes in a graph allowing for long-range interactions among voxels. Consequently, the shrinking bias of grid-structured pairwise CRFs is reduced allowing for more detailed segmentation results. In two key papers, Kr\"{a}henb\"{u}hl et al. proposed an efficient inference \citep{Krahenbuhl2012} and learning algorithm \citep{Kraehenbuehl2013} for dense CRFs, making their application in medical image volumes feasible. As a consequence, dense CRFs are becoming increasingly popular for a wide range of segmentation problems in medical image analysis including brain tumor segmentation \citep{Kamnitsas2016}, lung segmentation \citep{Gao2016}, retinal vessel segmentation \citep{Orlando2014,Orlando2016} or liver segmentation \citep{Christ2016}. However, the assessment of segmentation uncertainty in dense CRFs has not been possible so far.

The need for uncertainty quantification in segmentation has recently been addressed by \citet{Le2016}, who proposed a Gaussian Process for sampling candidate segmentation boundaries close to a reference segmentation. Although applicable to multiple label maps simultaneously, their approach works on contours and is thus primarily designed for binary segmentation problems. In addition, the method is also decoupled from the segmentation method that generated the reference segmentation in the first place, and it is driven by the fuzzyness of image boundaries. Related to our work are also Bayesian approaches that rely on parameter sampling using MCMC techniques to quantify segmentation uncertainty (e.g. \citep{Iglesias2013,LeFolgoc2017}). In contrast to such methods, we do not aim to sample model parameter values but aim to rather perturb a fixed parameter value in a principled fashion. Recently, Papandreou et al. proposed random MAP perturbations (Perturb-and-MAP \citep{Papandreou2011}) for CRFs that effectively allow drawing samples from the underlying Gibbs distribution. \citet{Alberts2016} employed Perturb-and-MAP within a grid-structured CRF for segmenting brain tumors and quantifying the uncertainty of volume estimates. In contrast to \citep{Alberts2016,Le2016}, we focus on multi-label segmentation. More importantly, Perturb-and-MAP requires an exact MAP estimation, which is not feasible for dense CRFs in medical images due to their size. Moreover, in multi-label segmentation problems the MAP solution can usually only be approximated. Despite these obstacles, we propose a novel perturbation model, referred hereafter as Perturb-and-MPM, for dense multi-label CRFs and investigate its feasibility for quantifying segmentation uncertainty.

Recently, Kim et al. \citep{Kim2016} proposed a method that enables approximate sampling from CRFs for which MAP inference can be reformulated as an Integer Linear Program (ILP). Similarly to our approach, they also rely on random perturbations in order to realize the sampling. However, their approach requires to solve the respective ILP, which in case of a dense multi-label CRF would rely on a very large number of constraints and thus result in a running time prohibitive for clinical use.

The contribution of this paper is a novel approach to quantify segmentation uncertainty in dense multi-label CRFs, which is computationally efficient and easy to implement. The method is validated on synthetic data and on Magnetic Resonance (MR) Imaging datasets of brain tumor patients. A clinical application to the challenging problem of estimating the extent of resection and residual tumor volume in postoperative images is presented and serves as an example on how uncertainty estimation can improve the extraction of volumetric imaging biomarkers.
\section{Preliminaries}
This section introduces the notation and concepts this work is based on.
\label{section:Preliminaries}
\subsection{Notation} In the following, the input image is denoted by $I$. The set of voxels in $I$ is denoted by $V$ and the total number of voxels by $N$. The available label values are contained in the set \begin{math}\mathcal{L}=\left\{1,\ldots,m\right\}\end{math}. A labeling of $I$ is referred to by $X=\left\{x_{i}\right\}_{i\in V}$ with $x_{i}$ being a scalar value that indicates the segmented image region (e.g. tissue compartment), i.e. $x_{i}\in\mathcal{L}$.
\subsection{Conditional Random Field} In a CRF model, voxels of an image are represented by nodes in a graph. Hence, we interpret the image $I$ as an undirected graph $G$ consisting of nodes and edges. Every node in $G$ is associated with a random variable $x_{i}$. The pair $(I,X)$ is a CRF, if for any given $I$ the distribution $P\left(X|I\right)$ factorizes according to $G$. The conditional distribution corresponds to a Gibbs distribution $P\left(X|I\right)=\frac{1}{Z(I)}\exp{\left(-E(X|I)\right)}$, where $Z(I)$ is the partition function. The most probable (MAP) labeling for a given image $I$ can then be estimated by minimizing the energy, i.e.
\begin{equation}
\tilde{X}=\arg\min_{X}E(X|I)=\arg\min_{X}\left\{\sum_{i}\psi_{u}(x_{i}) + \sum_{i\sim j}\psi_{p}(x_{i},x_{j})\right\}.
\label{eq:MAPinference}
\end{equation}
The energy of a pairwise CRF corresponds to a sum of unary potentials $\psi_{u}$ (=data-likelihood) and pairwise potentials $\psi_{p}$ (=prior). In contrast to grid-structured CRFs, pairwise potentials in dense CRFs are defined between all pairs of voxels, i.e. $G$ is a complete graph.
\subsection{Dense CRF and MPM-inference}
\label{section:PreliminariesDenseCRF}
Kr\"{a}henb\"{u}hl et al. \citep{Krahenbuhl2012} proposed an efficient approximate Maximum Posterior Marginal (MPM) inference scheme based on a mean field approximation and cross bilateral filtering techniques. In short, a mean field with the approximate distribution $Q(X)=\prod_{i}Q_{i}(x_{i})$ is introduced. The mean field approximation algorithm aims at minimizing the KL-Divergence between the approximate distribution $Q$ and the true distribution $P$. The algorithm then iteratively performs updates of the marginals $Q_{i}(x_{i})$ via an efficient filtering scheme and estimates the MPM solution via $x_{MPM}=\arg\max_{l}Q_{i}(x_{i}=l)$ for all voxels in the image (where $l\in\mathcal{L}$). The update equation for a marginal $Q_{i}$ is given by:
\begin{equation}
Q_{i}(x_{i}) = \frac{1}{Z_{i}}\exp{\left(-\psi_{u}(x_{i})-\sum_{l^{'}\in\mathcal{L}}\sum_{j\neq i}Q_{j}(x_{j}=l^{'})\psi_{p}(x_{i},x_{j}=l^{'})\right)}
\label{eq:update}
\end{equation}
where for notational convenience we define the energy as
\begin{equation}
E_{Q}(x_{i};\theta) = \psi_{u}(x_{i})+\sum_{l^{'}\in\mathcal{L}}\sum_{j\neq i}Q_{j}(x_{j}=l^{'})\psi_{p}(x_{i},x_{j}=l^{'})
\label{eq:energy}
\end{equation}
and $\theta$ refers to both $\psi_{u}$ and $\psi_{p}$. The marginal probabilities of the original Gibbs distribution are approximated by: $P(x_{i}) \approx Q_{i}(x_{i}) = \frac{1}{Z_{i}}\exp{\left(-E_{Q}(x_{i};\theta)\right)}$.
\subsection{Perturb-and-MAP} The main idea of Perturb-and-MAP \citep{Papandreou2011} is to locally perturb potentials of the energy function of the CRF by adding independent and identically distributed (i.i.d.) random noise $\gamma$. Subsequently, the MAP labeling of the perturbed energy can be estimated as:
\begin{equation}
\tilde{X}=\arg\min_{X}\left\{E(X|I) + \gamma(X)\right\}.
\label{eq:perturbedMAPinference}
\end{equation}
If the perturbation density follows a Gumbel distribution, it can be shown that the Perturb-and-MAP model approximates the Gibbs distribution of the corresponding random field \citep{Tomczak2016}. More specifically, we consider $\gamma(X)$ to be drawn i.i.d. from a Gumbel distribution with zero mean and cumulative distribution function $F(t)=\exp(-\exp(-(t+c)))$, with $c$ being the Euler constant. Under perturbation of all potentials (cf. \citep{Papandreou2011}, Proposition 3), the MAP labeling of the perturbed energy coincides with the corresponding sample of the Gibbs distribution. In this case, the distribution of the perturbed MAP solutions (see Equation (\ref{eq:perturbedMAPinference})) is equivalent to the original Gibbs distribution:
\begin{equation}
P(X=\arg\min_{\hat{X}}\{E(\hat{X}|I) + \gamma(\hat{X})\})=\frac{1}{Z(I)}\exp{\left(-E(X|I)\right)}.
\label{eq:PerturbMAPmodel}
\end{equation}
We denote the LHS of Equation (\ref{eq:PerturbMAPmodel}) as Perturb-and-MAP model. This relationship is deeply connected to the so-called ``Gumbel-max-trick'' \citep{Yellott1977} which is summarized in the following Lemma \citep{Papandreou2011}:
\newtheorem{lem}{Lemma}
\begin{lem}
\label{lem:lemma1}
Let $\left\{\theta_{1},\ldots,\theta_{m}\right\}$, with $\theta_{j}\in\mathbb{R}$. We additively perturb them by $\tilde{\theta}_{j}=\theta_{j} + \gamma_{j}$ with $\gamma_{j}$ i.i.d. Gumbel samples. Then the probability that $\tilde{\theta}_{j}$ attains the minimum value is $P(\arg\min\{\tilde{\theta}_{1},\ldots,\tilde{\theta}_{m}\}=j)=\exp(-\theta_{j})/\sum_{j'=1}^{m}\exp(-\theta_{j'})$.
\end{lem}
\subsection{Estimation of marginal probability distributions}
\label{section:HistogramApprox}
We are interested in the marginal probability of a voxel $i$ being labeled with a particular label $l$, i.e. the event $x_{i}=l$ with $l\in\mathcal{L}$. The corresponding marginal probability distribution can be expressed as an expectation ($\mathbbm{1}\left\{\cdot\right\}$ denotes the indicator function):
\begin{equation}
P(x_{i}=l)=\mathbb{E}\left[\mathbbm{1}\left\{x_{i}=l\right\}\right].
\end{equation}
The expectation can be estimated via sample approximations. Consider $\mathcal{S}=\left\{\tilde{x}_{i}^{(1)},\tilde{x}_{i}^{(2)},\cdots,\tilde{x}_{i}^{(m)}\right\}$ to be $\left|\mathcal{S}\right|$ samples of possible labels for voxel $i$. The marginal probability for $x_{i}=l$ can then simply be estimated by
\begin{equation}
P(x_{i}=l)\approx\frac{1}{\left|\mathcal{S}\right|}\sum_{s\in\mathcal{S}}\mathbbm{1}\left\{\tilde{x}_{i}^{(s)}=l\right\}.
\end{equation}
This corresponds to the approximation of $P(x_{i})$ with the relative frequencies of its events, i.e. via the respective histogram.
\subsection{Error measures}
In the following, we rely on the voxel-wise hamming loss for quantifying differences between two image labelings $\tilde{X}$ and $X$:
\begin{equation}
\ell(\tilde{X},X)=\frac{1}{\left|V\right|}\sum_{i=1}^{\left|V\right|}\mathbbm{1}\left\{\tilde{x}_{i}\neq x_{i}\right\}.
\label{eq:voxelHammingLoss}
\end{equation}
Moreover, for measuring the similarity between two probability distributions, we introduce the (voxel-wise) total variation distance:
\begin{equation}
\left\|P(X)-Q(X)\right\|_{1}=\frac{1}{\left|V\right|}\sum_{i=1}^{\left|V\right|}\frac{1}{2}\sum_{l\in\mathcal{L}}\left|P(x_{i}=l)-Q(x_{i}=l)\right|.
\end{equation}
\section{Perturb-and-MPM}
\label{section:Methods}
In this section, we introduce the main contribution of our work, which is a perturbation-based sampling approach for dense multi-label CRFs. We start by employing a random field over the random variables \begin{math}X=\left\{x_{1},\ldots,x_{N}\right\}\end{math} conditioned on an image $I$, where every voxel $i=1,\ldots,N$ is associated with a random variable $x_{i}$. The random variables take values from the previously defined label set $\mathcal{L}=\left\{1,2,\ldots,m\right\}$. As presented in Section \ref{section:Preliminaries}, the dense CRF defines a Gibbs distribution $P\left(X|I\right)$ with energy $E\left(X|I\right)$ composed of unary potentials $\psi_{u}(x_{i})$ and pairwise potentials $\psi_{p}(x_{i},x_{j})$, which are collectively referred to as $\theta$.
\subsection{Perturbation Model}
\label{section:PerturbationModel}
\begin{figure}
\includegraphics[width=\textwidth]{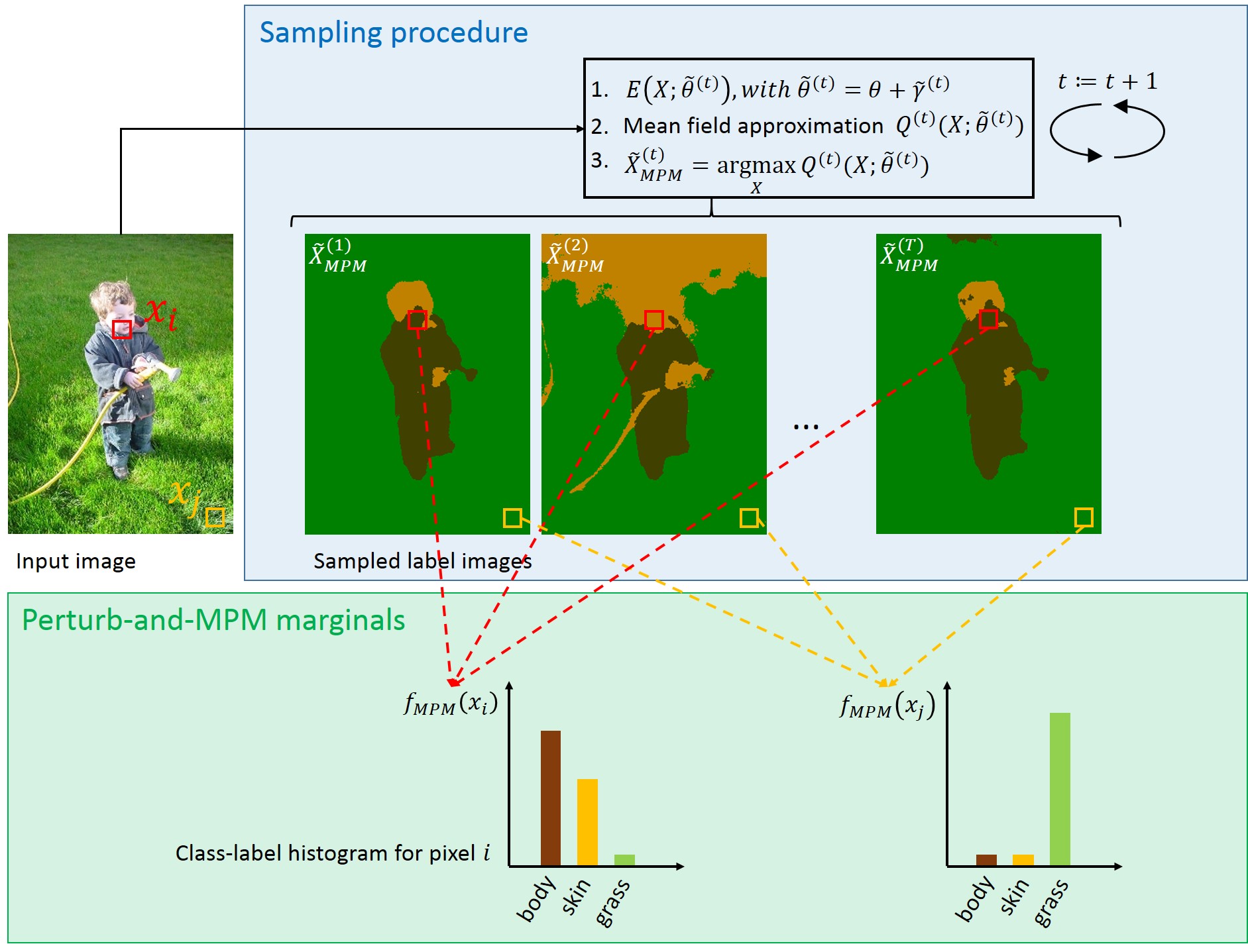}
\caption{Conceptual overview of Perturb-and-MPM. The input to Perturb-and-MPM is an image represented by a dense CRF and its energy function $E(X|I)$, respectively. \textbf{Sampling} is realized by first perturbing the potentials $\theta$. Second, by computing the mean field approximation $Q(X;\tilde{\theta})$ and third by estimating the MPM solution of the perturbed mean field. After iterating these three steps, the sampled labelmaps $\tilde{X}_{MPM}^{(j)}$ can be employed to estimate a class-label histogram for every pixel/voxel $i$ in the image. The class-label histogram serves as an estimate of the \textbf{Perturb-and-MPM marginals} $f_{MPM}(x_{i})$, which in turn are an approximation of the true but unknown marginals $P(x_{i})$ of the Gibbs-distribution. The colors in the histogram follow the colors of the labels in the segmented image. (Best seen in colors).}
\label{fig:2dexample}
\end{figure}

In the following, we adapt the perturbation strategy proposed by Papandreou et al. \citep{Papandreou2011}, to enable sampling from dense multi-label CRFs. The adaptation considers three steps: First, the potentials $\theta$ of the initial energy function are perturbed with i.i.d. Gumbel noise $\gamma$. Second, the mean field approximation is performed. Third, the perturbed (voxel-wise) MPM-problem is solved, i.e. $\tilde{x}_{MPM,i}=\arg\max_{l}\left\{Q_{i}(x_{i}=l;\tilde{\theta})\right\}$ with $\tilde{\theta}=\theta + \gamma$. The Perturb-and-MPM model $f_{MPM}$ can be constructed by iterating the previous three steps and aggregating the MPM solutions. This enables the estimation of the marginal distribution $f_{MPM}(x_{i})$ via the histogram of the aggregated MPM solutions (see Section \ref{section:HistogramApprox}). The maximum number of samples (i.e. aggregated MPM solutions) is denoted by $T$. Note that $f_{MPM}(x_{i}=l)$ is used as a shorthand for $P(\tilde{x}_{MPM,i}=l)$ and they can be used interchangeably. The main idea is visualized in Figure \ref{fig:2dexample} and the procedure is summarized in Algorithm \ref{listing:Perturb-and-MPM}.
\begin{algorithm}
\caption{Perturb-and-MPM}
\label{listing:Perturb-and-MPM}
\begin{algorithmic}[1]
 \State Sample set $\mathcal{S}=\left\{\emptyset\right\}, t := 0$, Maximum number of samples $T$
 \While{$(t<T)$}
	 \State Sample $\gamma^{(t)}$ from zero mean Gumbel distribution
	 \State Perturb potentials of $E(X|I;\theta)$ $\rightarrow\tilde{\theta}^{(t)}:=\theta+\gamma^{(t)}$
   \State Perform mean field approximation of dense CRF $\rightarrow Q^{(t)}(X|I;\tilde{\theta}^{(t)})$ \citep{Krahenbuhl2012}
    \State Solve $\tilde{X}_{MPM}^{(t)}=\arg\max_{X}\left\{Q^{(t)}(X|I;\tilde{\theta}^{(t)})\right\}$
		\State $\mathcal{S}^{(t+1)} := \mathcal{S}^{(t)}\cup\left\{\tilde{X}_{MPM}^{(t)}\right\}$
    \State $t := t + 1$
\EndWhile
\State Estimate $f_{MPM}$ using $\mathcal{S}$
\end{algorithmic}
\end{algorithm}
\subsection{Distribution of MPM-solutions}
\label{section:DistributionMPM}
\begin{figure}
\includegraphics[width=\textwidth]{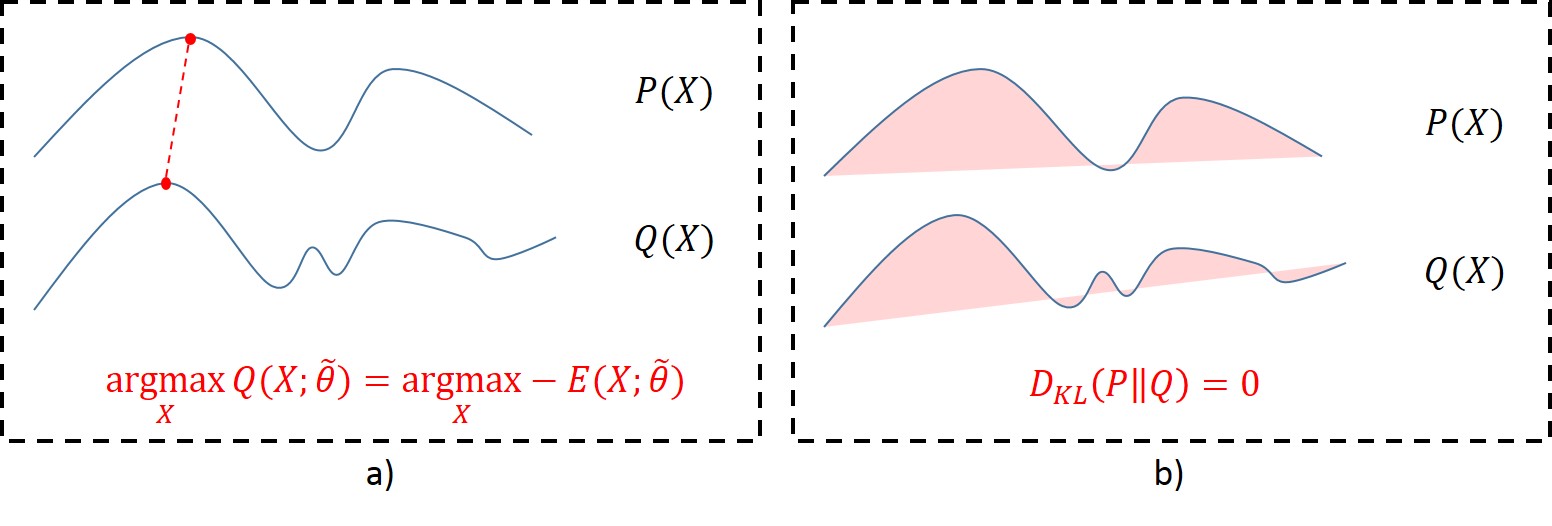}
\caption{Intuition of the proposed perturbation model. Perturb-and-MPM enables sampling from the original Gibbs distribution $P(X)$ if the MPM and MAP-solutions coincide (a, maxima indicated in red). This requirement of coinciding maxima is less restrictive than a mean field approximation of $P(X)$ (b).}
\label{fig:maxima}
\end{figure}

A natural question that arises from the previous perturbation model is how the perturbed MPM solutions are distributed. This aspect is crucial since our estimation of segmentation uncertainty will be based on this distribution. Given $\mathcal{L}$, one out of $m$ label values can be assigned to a variable $x_{i}$. We proceed by reparametrizing Equation (\ref{eq:energy}) to explicitly take all possible label assignments into account (cf. fully expanded potential table of \citep{Papandreou2011}): 
\begin{equation}
\bar{E}_{Q}(x_{i},\bar{\theta})=\left\langle \bar{\theta},\bar{\Psi}(x_{i})\right\rangle=E_{Q}(x_{i};\theta)
\label{eq:reparam}
\end{equation}
with $\bar{\theta}=\left[E_{Q}(x_{i}=1;\theta),\ldots,E_{Q}(x_{i}=m;\theta)\right]^{T}$, $\bar{\Psi}=\left[\mathbbm{1}\left\{x_{i}=1\right\},\ldots,\mathbbm{1}\left\{x_{i}=m\right\}\right]^{T}$.
Now, we can state the following proposition:
\begin{prop}
\label{prop:proposition1}
If we perturb each entry of the fully expanded potential vector $\bar{\theta}$ with i.i.d. Gumbel noise samples $\gamma_{j}$, $j=1,\cdots,m$, then the marginals of the Perturb-and-MPM and mean field model coincide, i.e., $P(\tilde{x}_{MPM,i} = x)=Q_{i}(x).$
\end{prop}
\begin{pf}
The proof is analogous to the proof of Proposition 3 in \citep{Papandreou2011} and makes use of the reparametrization in Equation (\ref{eq:reparam}) and Lemma \ref{lem:lemma1}.
\begin{align}
P(\arg\min_{x}\left\{\bar{E}_{Q}(x_{i}=x;\bar{\theta})\right\} = l)&=& P(\arg\min\left\{\bar{\theta}_{1}+\gamma_{1},\ldots,\bar{\theta}_{m}+\gamma_{m}\right\} = l) \\
&=& \frac{\exp\left(-\bar{\theta}_{l}\right)}{\sum_{l^{'}=1}^{m}\exp{\left(-\bar{\theta}_{l^{'}}\right)}} \\
&=& \frac{\exp\left(-E_{Q}(x_{i}=l;\theta)\right)}{\sum_{l^{'}=1}^{m}\exp{\left(-E_{Q}(x_{i}=l^{'};\theta\right)}} \\
&=& \frac{1}{Z_{i}}\exp\left(-E_{Q}(x_{i}=l;\theta)\right) = Q_{i}(x_{i}=l)
\end{align}
\qed
\end{pf}
The consequence of Proposition \ref{prop:proposition1} is that the Perturb-and-MPM model is equivalent to the original mean field approximation. However, Proposition \ref{prop:proposition1} does not hold for the Algorithm \ref{listing:Perturb-and-MPM}, because the mean field approximation $Q_{i}(x)$ will be different in every iteration (i.e. for every perturbation, cf. step 4\&5 in Algorithm \ref{listing:Perturb-and-MPM}). This observation plays a central role in the proposed Perturb-and-MPM framework. It allows the generation of marginal distributions that are possibly closer to the marginals of the original Gibbs distribution $P(x_{i})$ than mean field marginals $Q_{i}(x_{i})$. From Equation (\ref{eq:PerturbMAPmodel}), we know that the distribution of perturbed MAP-solutions is equivalent to the original Gibbs distribution, i.e. $P(\tilde{X}_{MAP}=X)=P(X)$ with $\tilde{X}_{MAP}=\arg\max_{X}-E(X;\tilde{\theta})$. If we assume that our MPM solution coincides with the MAP, i.e. $\arg\max_{X}Q(X;\tilde{\theta})=\arg\max_{X}-E(X;\tilde{\theta})$ holds, then we can state that $P(\tilde{X}_{MPM}=X)=P(\tilde{X}_{MAP}=X)=P(X)$. This requirement of coinciding maxima, illustrated in Figure \ref{fig:maxima}, is much less restrictive than requiring a perfect mean field approximation (i.e. KL-Divergence=0). Mean field approximations have shown to suffer from bad local optima \citep{Hoffman2015}. Thus, we anticipate that $f_{MPM}(X)$ may potentially yield marginals that are closer to the true distribution than the mean field approximation.
\subsection{Deriving error bounds of Perturb-and-MPM}
In the previous section, we demonstrated that the Perturb-and-MPM model can generate marginals different from the mean field marginals $Q_{i}(x_{i})$ and hence possibly better approximate the marginals of the original Gibbs distribution than a mean field approximation could do. The aim of this section is to perform an error analysis on the derived Perturb-and-MPM model. We are interested in obtaining bounds on the approximation error of Perturb-and-MPM, as this serves as a basis for deriving additional bounds on the error of the final uncertainty quantification. The Perturb-and-MAP and Perturb-and-MPM solutions are given by $\tilde{X}_{MAP}=\arg\max_{X}-E(X;\tilde{\theta})$ and $\tilde{X}_{MPM}=\arg\max_{X}Q(X;\tilde{\theta})$, respectively. In order to effectively tackle the analysis, we differentiate between two different situations: i) $\tilde{X}_{MAP}=\tilde{X}_{MPM}$ and ii) $\tilde{X}_{MAP}\neq\tilde{X}_{MPM}$. The corresponding loss function is the voxel-wise hamming loss as defined in Equation (\ref{eq:voxelHammingLoss}). Furthermore, we rely on three different assumptions:
\begin{enumerate}
  \item Equal number of samples $\left|\mathcal{S}\right|$ for the Perturb-and-MAP $f_{MAP}$ and Perturb-and-MPM model $f_{MPM}$.
	\item Perturbation of all potentials (unary and pairwise) such that $P(\tilde{X}_{MAP}=X)=P(X)$ holds (cf. Proposition 3 in Papandreou et al. \citep{Papandreou2011}).
	\item Fixed model $\theta$.
\end{enumerate}
\subsubsection{Situation $\tilde{X}_{MAP}=\tilde{X}_{MPM}$}
First, we start by looking at the situation where $\tilde{X}_{MAP}=\tilde{X}_{MPM}$. The Perturb-and-MAP framework states that under perturbation of all potentials, the identity $P(X)=P(\tilde{X}_{MAP}=X)$ holds. We have $f_{MPM}(X)=P(\tilde{X}_{MPM}=X)$ with $\tilde{X}_{MPM}=\arg\max_{X}Q(X;\tilde{\theta})$. If we assume that our MPM solution coincides with the MAP ($\arg\max_{X}Q(X;\tilde{\theta})=\arg\max_{X}-E(X;\tilde{\theta})$), we can then state that $f_{MPM}(X)=P(\tilde{X}_{MAP}=X)=P(X)$. In other words, the Perturb-and-MPM model coincides with the original Gibbs distribution. The model $f_{MPM}(X)$ is estimated based on a finite number of samples $\left|\mathcal{S}\right|$. We are thus interested i) in knowing if the respective approximation error is bounded and ii) in how many samples are necessary to achieve this bound.

We define the marginals of the Perturb-and-MPM model to be $f_{MPM}(x_{i}=l)=\mathbb{E}\left[\mathbbm{1}\left\{\tilde{x}_{MPM,i}=l\right\}\right]$. As presented in Section \ref{section:HistogramApprox}, the expectation can be estimated via sample approximation:
\begin{equation}
f_{MPM}(x_{i}=l)\approx\frac{1}{\left|\mathcal{S}\right|}\sum_{s\in\mathcal{S}}\mathbbm{1}\left\{\tilde{x}_{MPM,i}^{(s)}=l\right\}.
\label{eq:fMPMsampleApprox}
\end{equation}
Note that these voxel-wise marginals are computed independently from each other as the mean over a set of samples $\mathcal{S}$. We denote the previous sample approximation by $\hat{f}_{MPM}(x_{i}=l)$. This sum approximates the true but unknown Perturb-and-MPM marginal $f_{MPM}$. Hence, we refer to it as the empirical Perturb-and-MPM model. Furthermore, $\mathbbm{1}\left\{\tilde{x}_{MPM,i}=l\right\}$ is a Bernoulli random variable and thus the estimate $\hat{f}_{MPM}(x_{i}=l)$ defines $\left|\mathcal{S}\right|$ independent Bernoulli trials, each with success probability $f_{MPM}(x_{i}=l)$. Consequently, we can bound the absolute error between the empirical and the true but unknown Perturb-and-MPM marginal via the Hoeffding bound:
\begin{equation}
P\left(\left|\hat{f}_{MPM}(x_{i}=l)-f_{MPM}(x_{i}=l)\right|\geq\epsilon\right)\leq 2\exp(-2\left|\mathcal{S}\right|\epsilon^{2})
\label{eq:primitiveBound}
\end{equation}
If we choose $\delta$ such that
\begin{equation}
2\exp(-2\left|\mathcal{S}\right|\epsilon^{2})\leq\delta,
\end{equation}
we can say that with a probability of at least $1-\delta$ it holds that:
\begin{equation}
\left|\hat{f}_{MPM}(x_{i}=l)-f_{MPM}(x_{i}=l)\right|\leq\epsilon.
\end{equation}
In this situation (given $\tilde{X}_{MAP}=\tilde{X}_{MPM}$ holds) the approximation error of the empirical Perturb-and-MPM model with respect to the original Gibbs distribution is bounded:
\begin{align}
&\left|\hat{f}_{MPM}(x_{i}=l)-f_{MPM}(x_{i}=l)\right| \\
&=\left|\frac{1}{\left|\mathcal{S}\right|}\sum_{s\in\mathcal{S}}\mathbbm{1}\left\{\tilde{x}_{MPM,i}^{(s)}=l\right\}-P(x_{i}=l)\right| \\
&=d\leq\epsilon
\label{eq:bound1}
\end{align}
The first equality follows from the previous reasoning that if $\tilde{X}_{MAP}=\tilde{X}_{MPM}\Longrightarrow f_{MPM}(X)=P(X)$ and from the definition of $\hat{f}_{MPM}(x_{i}=l)$ provided by Equation (\ref{eq:fMPMsampleApprox}). The law of large numbers implies that with an increasing number of samples $\left|\mathcal{S}\right|$ the approximation error $d$ shrinks to zero, $\lim\limits_{\left|\mathcal{S}\right| \to \infty}d=0$. In other words, with a sufficiently large number of samples the empirical Perturb-and-MPM marginal coincides with the marginal of the original Gibbs distribution. Note that this holds for any voxel $i$ in the image. For fixed $(\epsilon,\delta)$ the required sample size to obtain a reliable Perturb-and-MPM model is then given by:
\begin{equation}
\left|\mathcal{S}\right|\geq\frac{\log(2/\delta)}{2\epsilon^{2}}
\label{eq:primitiveSampleSize}
\end{equation}
The previous bound in Equation (\ref{eq:primitiveBound}) is limited since it holds only for a fixed label $l$. Thus, we generalize it in order to know how likely it is that any of the empirical marginals $\hat{f}_{MPM}(x_{i})$ significantly deviates from $P(x_{i})$. This can be described by the probability
\begin{equation}
\resizebox{.9\hsize}{!}{$P\left(\exists l \in \mathcal{L} : \left|\hat{f}_{MPM}(x_{i}=l)-f_{MPM}(x_{i}=l)\right|\geq\epsilon\right)=P\left(\max_{l \in \mathcal{L}}\left|\hat{f}_{MPM}(x_{i}=l)-f_{MPM}(x_{i}=l)\right|\geq\epsilon\right)$.}
\end{equation}
We proceed as follows:
\begin{align}
&P\left(\max_{l \in \mathcal{L}}\left|\hat{f}_{MPM}(x_{i}=l)-f_{MPM}(x_{i}=l)\right|\geq\epsilon\right) \\
&=P\left(\bigcup_{l \in \mathcal{L}}\left|\hat{f}_{MPM}(x_{i}=l)-f_{MPM}(x_{i}=l)\right|\geq\epsilon\right) \\
&\leq\sum_{l \in \mathcal{L}}P\left(\left|\hat{f}_{MPM}(x_{i}=l)-f_{MPM}(x_{i}=l)\right|\geq\epsilon\right) \\
&\leq\sum_{l \in \mathcal{L}}P\left(\left|\hat{f}_{MPM}(x_{i}=l)-P(x_{i}=l)\right|\geq\epsilon\right) \\
&\leq\sum_{l \in \mathcal{L}}2\exp(-2\left|\mathcal{S}\right|\epsilon^{2})=2\left|\mathcal{L}\right|\exp(-2\left|\mathcal{S}\right|\epsilon^{2}).
\end{align}
Based on this, we can refine the required sample size in Equation (\ref{eq:primitiveSampleSize}) and obtain
\begin{equation}
\left|\mathcal{S}\right|\geq\frac{\log(2\left|\mathcal{L}\right|/\delta)}{2\epsilon^{2}}.
\end{equation}
Therefore, the required number of samples $\left|\mathcal{S}\right|$ to obtain a reliable (in the sense of $(\epsilon,\delta)$) Perturb-and-MPM model, grows logarithmically with the number of labels $\left|\mathcal{L}\right|$.
\subsubsection{Situation $\tilde{X}_{MAP}\neq\tilde{X}_{MPM}$}
Second, we are looking at the situation where $\tilde{X}_{MAP}\neq\tilde{X}_{MPM}$. In this situation, the previously derived approximation error (Equation (\ref{eq:bound1})) is not guaranteed to shrink to zero anymore but to a residual error. We can characterize this error by investigating the loss that is induced through the different labelings $\tilde{X}_{MAP}\neq\tilde{X}_{MPM}$. We can state the following Proposition:
\begin{prop}
Given $\tilde{X}_{MAP}$ and $\tilde{X}_{MPM}$ with $\tilde{X}_{MAP}\neq\tilde{X}_{MPM}$ an upper bound on the hamming loss
\begin{equation}
0\leq\ell(\tilde{X}_{MPM},\tilde{X}_{MAP})\leq\epsilon
\end{equation}
implies that
\begin{equation}
0\leq\left\|P(X)-f_{MPM}(X)\right\|_{1}\leq\epsilon
\label{eq:bound}
\end{equation}
with $\left\|\cdot\right\|_{1}$ being the (voxel-wise) total variation distance.
\end{prop}
\begin{pf}
We assume for the Perturb-and-MAP ($f_{MAP}$) and Perturb-and-MPM ($f_{MPM}$) model that for both models the same number of samples $\left|S\right|$ was generated. Furthermore, we assume a perturbation of all potentials such that $P(X)=f_{MAP}(X)$ holds.
\begin{align}
&\left\|P(X)-f_{MPM}(X)\right\|_{1}=\left\|f_{MAP}(X)-f_{MPM}(X)\right\|_{1} \\
&=\frac{1}{\left|V\right|}\sum_{i=1}^{\left|V\right|}\frac{1}{2}\sum_{l\in\mathcal{L}}\left|f_{MAP}(x_{i}=l)-f_{MPM}(x_{i}=l)\right| \\
&\approx\frac{1}{2\left|V\right|} \sum_{i=1}^{\left|V\right|}\sum_{l\in\mathcal{L}}\left|\frac{1}{\left|\mathcal{S}\right|}\sum_{s\in\mathcal{S}}\mathbbm{1}\left\{\tilde{x}_{MAP,i}^{(s)}=l\right\} - \frac{1}{\left|\mathcal{S}\right|}\sum_{s\in\mathcal{S}}\mathbbm{1}\left\{\tilde{x}_{MPM,i}^{(s)}=l\right\}\right| \\
&\leq\frac{1}{2\left|V\right|} \sum_{i=1}^{\left|V\right|}\sum_{l\in\mathcal{L}}\frac{1}{\left|\mathcal{S}\right|}\sum_{s\in\mathcal{S}}\left|\mathbbm{1}\left\{\tilde{x}_{MAP,i}^{(s)}=l\right\} - \mathbbm{1}\left\{\tilde{x}_{MPM,i}^{(s)}=l\right\}\right| \\
&=\frac{1}{\left|\mathcal{S}\right|}\sum_{s\in\mathcal{S}}\frac{1}{2\left|V\right|}\sum_{i=1}^{\left|V\right|}\sum_{l\in\mathcal{L}}\left|\mathbbm{1}\left\{\tilde{x}_{MAP,i}^{(s)}=l\right\} - \mathbbm{1}\left\{\tilde{x}_{MPM,i}^{(s)}=l\right\}\right| \\
&=\frac{1}{\left|\mathcal{S}\right|}\sum_{s\in\mathcal{S}}\frac{1}{\left|V\right|}\sum_{i=1}^{\left|V\right|}\mathbbm{1}\left\{\tilde{x}_{MAP,i}^{(s)}\neq \tilde{x}_{MPM,i}^{(s)}\right\} \\
&=\frac{1}{\left|\mathcal{S}\right|}\sum_{s\in\mathcal{S}}\ell(\tilde{X}_{MAP}^{(s)},\tilde{X}_{MPM}^{(s)}) \\
&\leq\frac{1}{\left|\mathcal{S}\right|}\sum_{s\in\mathcal{S}}\epsilon = \frac{1}{\left|\mathcal{S}\right|}\left|\mathcal{S}\right|\epsilon = \epsilon
\end{align}
\qed
\end{pf}
This shows that the approximation error is indeed bounded. For $\epsilon<\left|V\right|$ the bound is tighter than the trivial $0\leq\left\|P(X)-\hat{f}_{MPM}(X)\right\|_{1}\leq\left|V\right|$. As a consequence, the voxel-wise error can be bounded in the same fashion $0\leq\left\|P(x_{i})-\hat{f}_{MPM}(x_{i})\right\|_{1}\leq\epsilon_{i}$ with $\epsilon_{i}=1$ being the worst case.

In summary, we found for the situation where $\tilde{X}_{MAP}=\tilde{X}_{MPM}$ holds:
\begin{itemize}
  \item The approximation error $\left\|P(X)-\hat{f}_{MPM}(X)\right\|_{1}$ is bounded and will go to zero with an increasing number of samples $\left|S\right|$.
	\item The required number of samples for a reliable Perturb-and-MPM model $f_{MPM}(X)$ scales logarithmically with the number of labels $\left|\mathcal{L}\right|$, i.e. $\left|\mathcal{S}\right|\propto\log(\left|\mathcal{L}\right|)$.
\end{itemize}
For the situation where $\tilde{X}_{MAP}\neq\tilde{X}_{MPM}$ holds, we found:
\begin{itemize}
  \item Given the hamming loss function $\ell(\tilde{X}_{MAP},\tilde{X}_{MPM})$, the approximation error $\left\|P(X)-f_{MPM}(X)\right\|_{1}$ can be bounded.
\end{itemize}

\subsection{Uncertainty Quantification}
\label{section:uncertaintyQuant}
Based on Algorithm \ref{listing:Perturb-and-MPM}, we can effectively generate the Perturb-and-MPM model $f_{MPM}$. The uncertainty of the segmentation model for a possible labeling $X$ can then be derived from the MPM-marginals $f_{MPM}(x_{i})$. The marginals reflect voxel-wise histograms. A natural choice to quantify uncertainty is via the Shannon-entropy contained in these histograms. The entropy for a discrete probability distribution $P$ over possible labels $l\in \mathcal{L}$ is given by: $H_{P}(x_{i})=-\sum_{l\in \mathcal{L}}P(x_{i}=l)\log_{2} P(x_{i}=l)$. The entropy is maximal for uniformly distributed labels. We can now use the previously derived bound in Equation (\ref{eq:bound}) and results from information theory (Equation (11) in Sason et al. \citep{Sason2013}) to provide an upper bound on the difference in entropy of the Perturb-and-MPM model and the original Gibbs distribution:
\begin{align}
&\left|H_{P}(x_{i}) - H_{f_{MPM}}(x_{i})\right| \\
&\leq\left\|P(x_{i})-f_{MPM}(x_{i})\right\|_{1} \log_{2}(\left|\mathcal{L}\right|-1) + h(\left\|P(x_{i})-f_{MPM}(x_{i})\right\|_{1}) \\
&\leq\epsilon_{i} \log_{2}(\left|\mathcal{L}\right|-1) + h(\epsilon_{i})
\end{align}
with $h$ denoting the binary entropy function. For the worst case scenario of $\epsilon_{i}=1$ the bound reduces to:
\begin{equation}
\left|H_{P}(x_{i}) - H_{f_{MPM}}(x_{i})\right|\leq\log_{2}\left(\left|\mathcal{L}\right|-1\right)
\label{eq:uncertaintyBound}
\end{equation}
As an example, if we set $\left|\mathcal{L}\right|=4$, we obtain the value $1.585$ for the RHS in Equation (\ref{eq:uncertaintyBound}). Hence, this bound is tighter than the trivial bound $\left|H_{P}(x_{i}) - H_{f_{MPM}}(x_{i})\right|\leq\log_{2}(\left|\mathcal{L}\right|)$, which in this case is $\log_{2}(\left|\mathcal{L}\right|=4)=2$.

\subsection{Implementation of Perturb-and-MPM}
Sampling from the zero mean, unit variance Gumbel distribution can be realized via drawing samples $u$ from the standard uniform distribution $U(0,1)$ and transforming them according to $-\log(-\log(u))$. Note that the number of perturbations for perturbing all potentials of the initial energy function grows exponentially with the number of nodes $N$ in the CRF. Recent studies \citep{Hazan2013,Papandreou2011} have empirically shown that it is sufficient to perform low-order Gumbel perturbations rather than perturbing all realizations of both unary and pairwise potentials. Hence, we employ an Order-1 perturbation which perturbs each of the unary potentials with Gumbel noise.
\section{Experimental results \& Application}
\subsection{Validation on synthetic data}
\begin{figure}
\includegraphics[width=\textwidth]{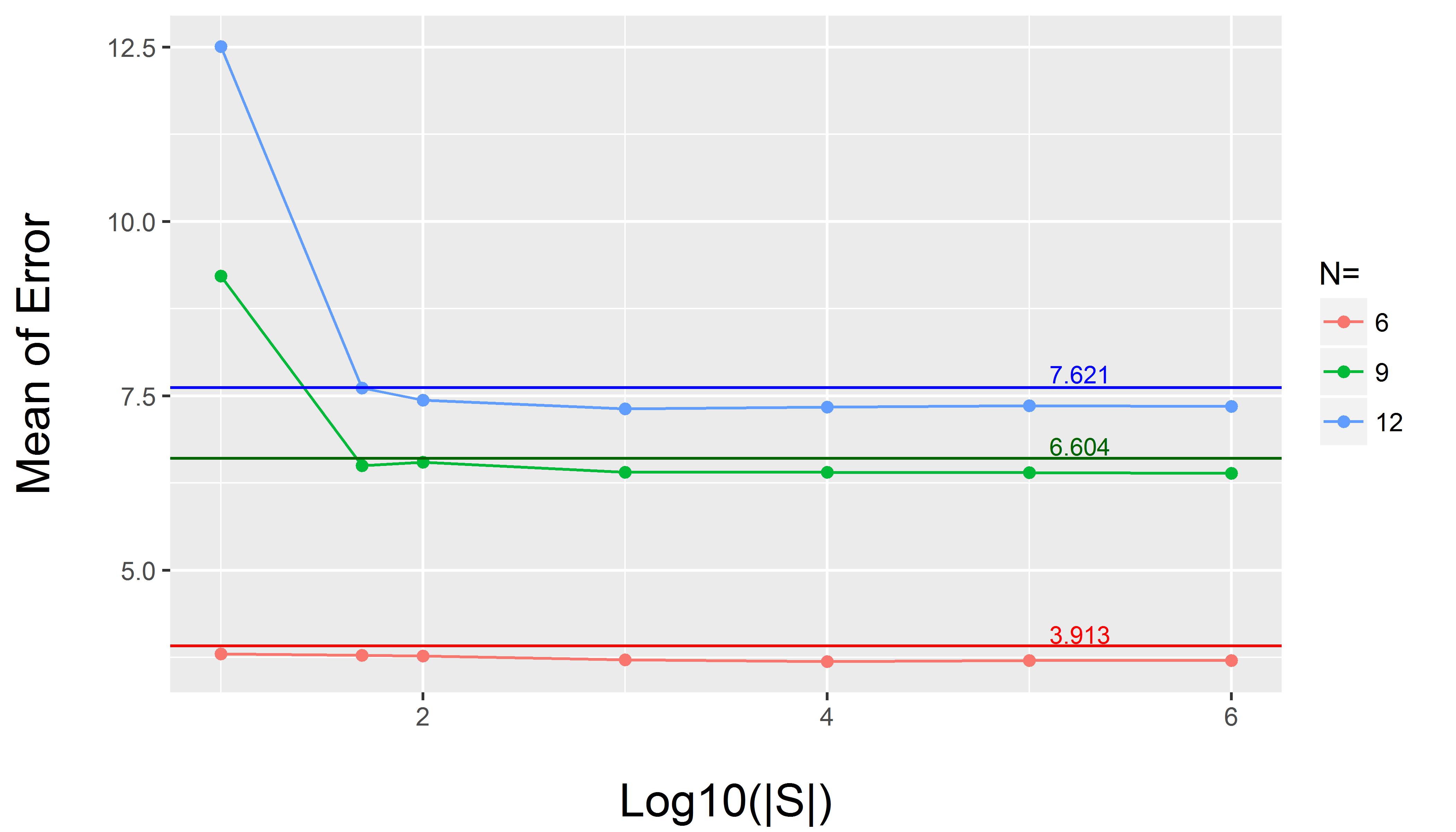}
\caption{Mean of absolute error of log-probabilities on synthetic data. For different number of samples $f_{MPM}$ marginals were estimated and their distance to the exact marginals was quantified using the $\ell_{1}$-norm. The mean was computed over 20 random initializations (of the unary potentials) for each measurement point. The constant lines indicate the $\ell_{1}$-distance between the unperturbed mean field marginals $Q(x_{i})$ and the exact marginals. (Best seen in colors).}
\label{fig:curveMeans}
\end{figure}

In order to empirically validate the proposed Perturb-and-MPM scheme, we employed a dense CRF defined over grids with different number of nodes ($N\in\left\{6,9,12\right\}$) which can take on binary labels, i.e. $x_{i}\in\left\{0,1\right\}$. The unary potentials were defined to be $\psi_{u}=-log(p_{u})$, where $p_{u}(x_{i}=0)\sim U[0,1)$ and $p_{u}(x_{i}=1) = 1-p_{u}(x_{i}=0)$. The pairwise potentials contained a Potts prior and a gaussian kernel over the node positions. For every grid size, we estimated $f_{MPM}$ with a varying number of samples $(\left|\mathcal{S}\right|\in\left\{10,50,10^{2},10^{3},10^{4},10^{5},10^{6}\right\}$. We performed a comparison of unperturbed mean field marginals $Q(x_{i})$ with marginals of $f_{MPM}$ and exact marginals (estimated with an exact sampler of the R package ``CRF'' \citep{CRFpackage}) over 20 random initializations of $\psi_{u}$ (a larger number of initializations did not change the result). Similarity of the unperturbed mean field $Q(x_{i})$ and $f_{MPM}$ marginals with exact marginals was quantified using the $\ell_{1}$-norm. The corresponding error plot is shown in Figure \ref{fig:curveMeans}. For an increasing number of samples $\left|S\right|$ the mean (over the 20 initializations) of the approximation error converges to a residual error and from $\left|S\right|=50$ on appears to be consistently lower than the mean approximation error of the unperturbed mean field (clearly visible for $N=9$, $N=12$).
\subsection{Assessing multi-label segmentation uncertainty in glioblastoma patient images}
For evaluating the clinical utility of our uncertainty estimation, we looked at the task of multi-label segmentation of glioblastoma from multi-sequential Magnetic Resonance Imaging (MRI) data. In a first application, we look at the most challenging subtask in glioblastoma segmentation, which is the segmentation of the tumor core, including the identification of necrosis, non-enhancing and enhancing tumor. The differentiation of edema from non-enhancing and enhancing tumor is connected to a high uncertainty for human raters, which in turn is reflected in the rather poor performance of supervised learning-based segmentation methods for this task \citep{Menze2015}. In this application, we aim to quantify the quality of the uncertainty prediction. In a second application, we are computing the extent of resection and residual tumor volume from the segmentation results of pre- and postoperative images. The extent of resection as well as the residual tumor volume are important clinically-relevant indicators of success of the neurosurgical tumor resection, as well as prognostic biomarkers in glioblastoma patients \citep{Grabowski2014}. The aim is to study the use of uncertainty quantification in order to improve their extraction. The dataset used for the first application contains preoperative images of 14 patients, while the second application contains pre- as well as immediate postoperative images (acquired within 72 hours after surgery) of 19 patients. All cases are newly diagnosed and histologically confirmed glioblastoma. For the second dataset of 19 patients, nine patients underwent subtotal extirpations/partial resection of enhancing tumor (PRET) and ten patients underwent complete resections of enhancing tumor (CRET).

For both datasets the Magnetic Resonance protocol encompassed $T_{1}$-weighted, $T_{1}$-weighted gadolinium-enhanced, $T_{2}$-weighted and FLAIR-weighted sequences. The FLAIR-sequence was an axial acquisition with an anisotropic voxel size of $1 mm\times1 mm\times3 mm$, whereas the remaining sequences were sagittal acquisitions with an isotropic voxel size of $1 mm\times1 mm\times1 mm$. The field of view for all sequences was $256\times256 mm^{2}$. Manual segmentation of all tumor tissues for the first dataset was performed by one expert rater (Neuroradiologist with more than six years of experience). For the second dataset, segmentation of the enhancing tumor was performed by four expert raters. These manual segmentations were used as groundtruth data for the estimation of the extent of resection. More details on the MR acquisition, the segmentation protocol as well as the inter-observer variability of the individual raters can be found in a related clinical study \citep{Meier2016c}.

For performing the experiments, we employed a recently proposed and clinically-evaluated segmentation method (\citep{Meier2016b, Meier2016}). The method relies on a decision forest classifier for definition of unary potentials in a dense CRF. Details on the feature extraction, decision forest and CRF can be found in \citep{Meier2014b,Meier2016b}. Parameters of the CRF were learned by minimizing the Intersection-Over-Union loss according to \citep{Kraehenbuehl2013}. We remark that the training dataset of this model is an independent one from the datasets used for evaluation here, and encompassed 54 patient cases, described in more detail in \citep{Meier2016}.
\subsubsection{Visualization of segmentation uncertainty}
\begin{figure}
\includegraphics[width=\textwidth]{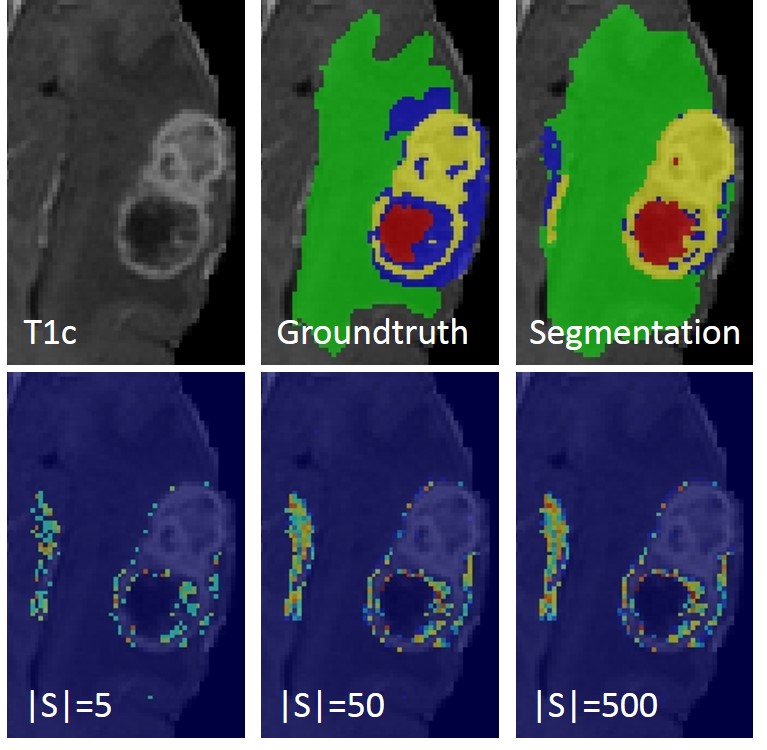}
\caption{Visualization of voxel-wise uncertainty with increasing number of samples. The segmentation of the tumor of the corresponding CRF model and the manual ground truth are shown. The individual tissues are necrosis (red), edema (green), enhancing (yellow) and non-enhancing (blue) tumor. The uncertainty is shown as jet heatmap overlay. (Best seen in colors).}
\label{fig:NumberOfSamples}
\end{figure}

The segmentation uncertainty is estimated as explained in Section \ref{section:uncertaintyQuant} and can be overlayed on the corresponding patient image in form of a heatmap, where voxel-wise uncertainty values are color-coded. In Figure \ref{fig:NumberOfSamples}, the impact of an increasing number of samples to estimate $f_{MPM}$ and subsequently the uncertainty is visualized. With an increasing number of samples the segmentation uncertainty estimation appears spatially more homogenous. We observed that the uncertainty estimation starts to saturate from $\left|\mathcal{S}\right|=100$ on (implications of these results are discussed in detail in Section \ref{section:discussion}).
\subsubsection{Quantifying the quality of the segmentation uncertainty in multi-label tumor core segmentation}
This experiment concerned the segmentation of the tumor core, which involves the differentiation of necrosis, enhancing and non-enhancing tumor from the surrounding edema. The aim of the experiment was to quantify the quality of the uncertainty estimation. In order to do so, we performed a comparison between the predicted segmentation, uncertainty and the manual ground truth data. Four situations were of interest:
\begin{itemize}
	\item Voxels that are misclassified \textbf{and} uncertain (=True Positives).
	\item Voxels that are misclassified \textbf{and} certain (=False Negative).
	\item Voxels that are correctly classified \textbf{and} uncertain (=False Positive).
	\item Voxels that are correctly classified \textbf{and} certain (=True Negative).
\end{itemize}
A total of $\left|\mathcal{S}\right|=200$ samples were generated in order to estimate segmentation uncertainty. For simplicity, we defined voxels to be uncertain if they showed a non-zero value in the uncertainty estimation. Sensitivity and specificity were assessed accordingly on the preoperative dataset containing 14 patients. The mean and standard deviation of the sensitivity and specificity (mean$\pm$sd) for the tumor core is $(0.45\pm0.23)$ and $(0.76\pm0.13)$. The values for the enhancing tumor are $(0.53\pm0.1)$  and $(0.88\pm0.14)$, respectively. By introducing a threshold value for exclusion of voxels with small uncertainty values, we obtain better specificity with the cost of having a reduced sensitivity. As an example a threshold for the uncertainty of 0.1 yields for the tumor core a sensitivity and specificity of $(0.39\pm0.21)$ and $(0.81\pm0.12)$, respectively, and for the enhancing tumor $(0.44\pm0.09)$ and $(0.93\pm0.11)$. Likewise, larger threshold values resulted in an increased specificity but lowered sensitivity.
\subsection{Extraction of prognostic biomarkers in glioblastoma patients}
\begin{figure}
\includegraphics[width=\textwidth]{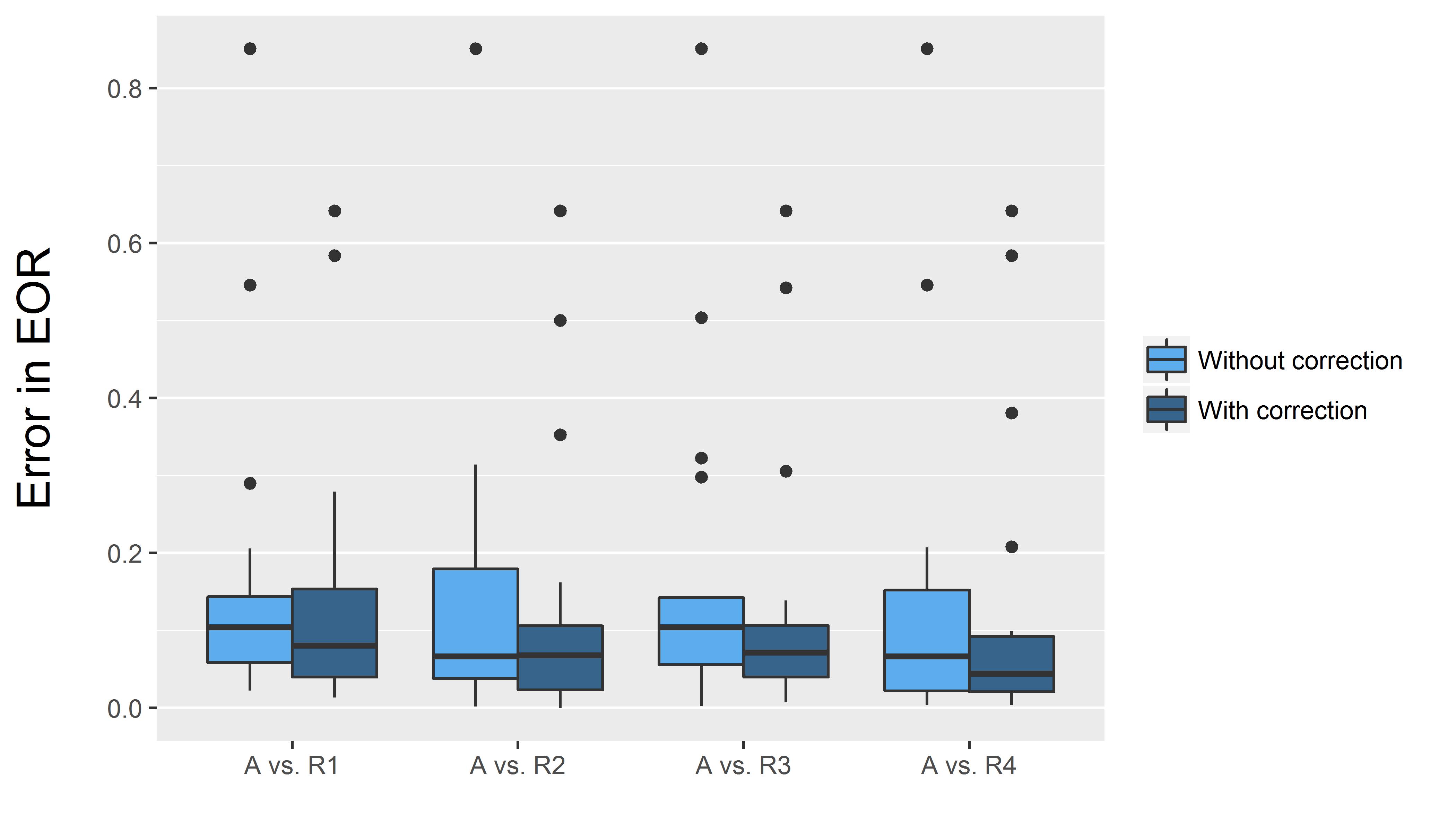}
\caption{Boxplot of absolute error in EOR. The error is measured between automatic estimates (A) and the estimates of each human rater (Rx). The light-colored box shows the error without correction, whereas the dark-colored box shows the error with correction. The dark line indicates the median.}
\label{fig:boxplotEOR}
\end{figure}

\begin{figure}
\includegraphics[width=\textwidth]{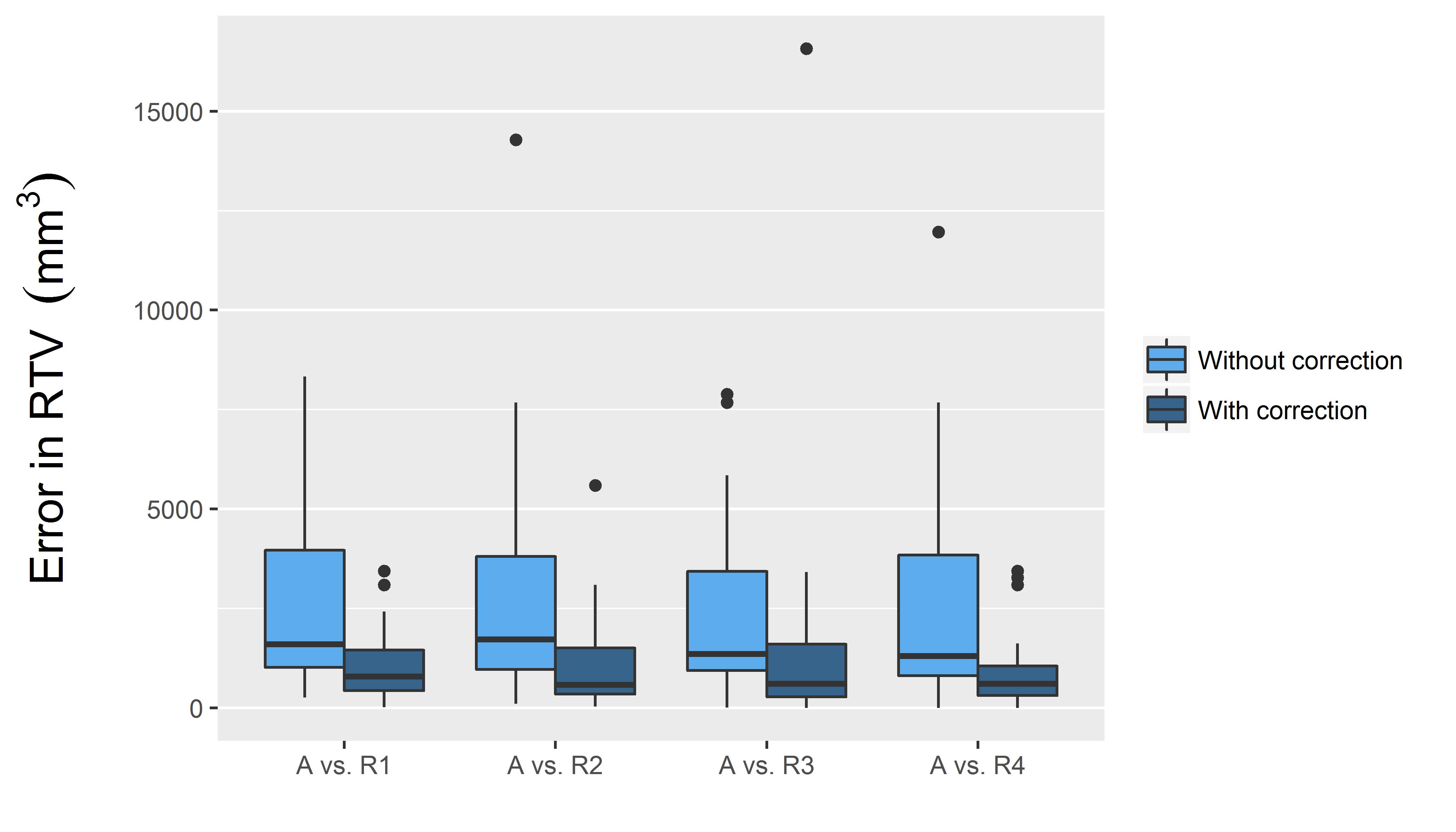}
\caption{Boxplot of absolute error in RTV. The error is measured between automatic estimates (A) and the estimates each human rater (Rx). The light-colored box shows the error without correction, whereas the dark-colored box shows the error with correction. The dark line indicates the median.}
\label{fig:boxplotRTV}
\end{figure}

\begin{figure}
\includegraphics[width=\textwidth]{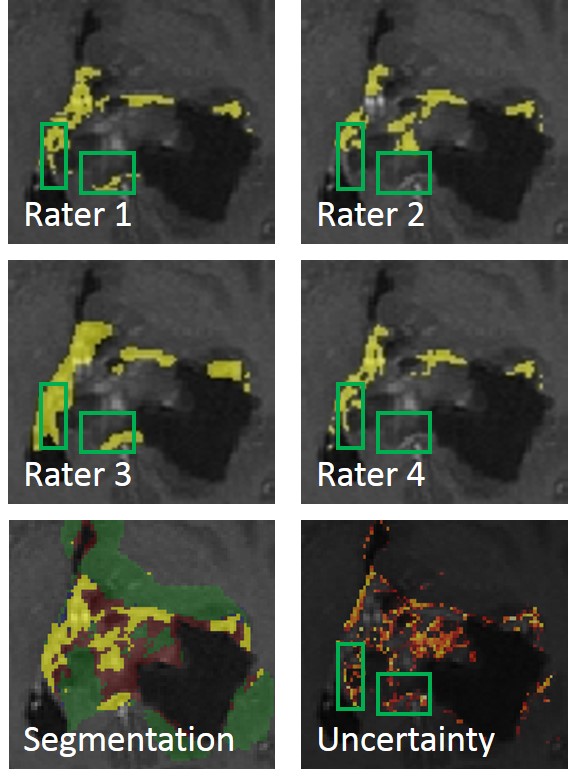}
\caption{Exemplary patient case with large residual enhancing tumor (=low EOR). Areas of large disagreement among expert raters (green boxes) correspond to regions of high uncertainty in the automatic segmentation model. The residual enhancing tumor is shown in yellow color. The segmentation of the hemorrhage is shown in red, edema is shown in green. The uncertainty is visualized as a hot heatmap with bright values indicating higher uncertainty. (Best seen in colors).}
\label{fig:comparisonRTV}
\end{figure}

\begin{figure}
\includegraphics[width=\textwidth]{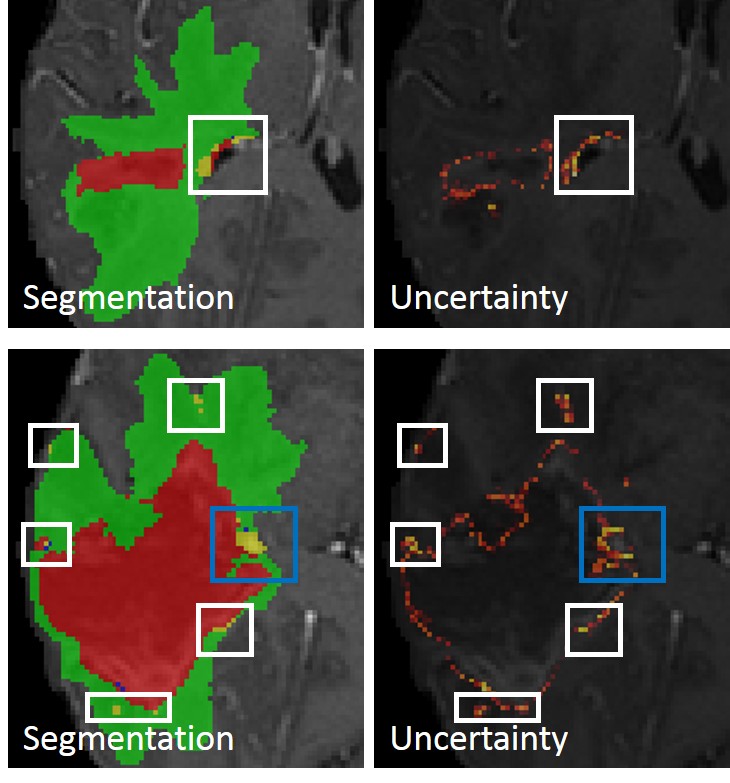}
\caption{Two exemplary slices of a patient with completely resected enhancing tumor (=high EOR). False positive segmentations of contrast-enhancing tumor (in yellow) associated with high uncertainty (white box) and partial uncertainty (blue box) are shown. The segmentation of the resection cavity/hemorrhage is shown in red, edema is shown in green. (Best seen in colors).}
\label{fig:visual-examples}
\end{figure}

In order to investigate the clinical utility of the uncertainty estimation, we looked at the challenging problem of estimating the extent of resection (EOR) and residual tumor volume (RTV). The extent of resection is defined by the preoperative enhancing tumor volume $V_{pre}$ and the residual enhancing tumor volume $V_{post}$ after surgery. It corresponds to the ratio:
\begin{equation}
EOR = \frac{V_{pre}-V_{post}}{V_{pre}}.
\label{eq:EOR}
\end{equation}
An EOR of 1 (or 100\%) reflects a complete resection of the enhancing tumor. Pre- and postoperative images of the second dataset containing 19 patients were segmented automatically by our approach \citep{Meier2016b}. The automatically estimated volumes of $V_{pre}$ and $V_{post}$ (=RTV) were used to compute EOR. The automatically estimated EOR and RTV were compared to the groundtruth data of four different raters. In a second step, we neglect all voxels that show a non-zero value in the uncertainty estimation for computing the EOR and RTV, respectively. $\left|\mathcal{S}\right|=200$ samples were generated to estimate segmentation uncertainty. The ``corrected'' EOR and RTV were then again compared to the groundtruth data. The resulting absolute error in EOR and RTV for both corrected and original estimation are presented as boxplots in Figures \ref{fig:boxplotEOR} and \ref{fig:boxplotRTV}. We performed a statistical analysis under a significance level of $\alpha=0.05$. The change in error between the original and corrected automatic estimates were assessed using a paired, exact Wilcoxon signed-rank test (R package ``exactRankTests'', Bonferroni correction was applied for multiple comparisons). We found a statistically significant decrease in error of the corrected automatic estimates for the RTV including the comparisons to rater 1 ($p=5.34\times10^{-5}$), rater 2 ($p=6.45\times10^{-4}$) and rater 4 ($p=1.64\times10^{-4}$). An exemplary uncertainty quantification of the segmentation for a postoperative case with a low EOR is shown in Figure \ref{fig:comparisonRTV}. Similarly, exemplary uncertainty maps for two different slices of a postoperative patient case with completely resected enhancing tumor (=high EOR) are shown in Figure \ref{fig:visual-examples}.
\section{Discussion \& Conclusion}
\label{section:discussion}
We have introduced a perturbation-based approach for performing efficient, approximate sampling in dense multi-label CRFs. The methodology enables the approximation of segmentation uncertainty contained in image labelings from dense CRFs. In medical image analysis, image segmentation is used to provide spatial information (volume/position) on anatomical regions of interest. In brain tumor image analysis, segmentation is used to partition glioblastoma into different tumor compartments. The utility of the spatial information of these tumor compartments is manifold. Possible areas of applications are longitudinal tumor volumetry \citep{Alberts2016,Meier2016}, planning and assessment of neurosurgical interventions \citep{Meier2016b,Porz2016} and radiomic analyses \citep{Cui2015,Gutman2015,RiosVelazquez2015}. We strongly believe that an estimate of segmentation uncertainty will help in any clinical decision-making process that is based on information from image segmentations. Uncertainty estimation could be used for various purposes such as the exclusion of uncertain labeled voxels from further analysis, or the guidance of an expert rater in the correction of automatically generated segmentations. The former situation would be of great importance for the emerging field of radiomics in which imaging biomarkers are discovered in a high-throughput setting based on features extracted from image segmentations \citep{Gillies2015,Yip2016}.
\subsection{Methodological aspects}
The proposed Perturb-and-MPM methodology enables approximate sampling from dense multi-label CRFs. We showed that the approximation error of Perturb-and-MPM is bounded and converges with an increasing number of samples to a residual error. For tractable problems, this residual error was shown to be consistently lower than the error of a mean field approximation. This empirically confirms our intuition in Section \ref{section:DistributionMPM} (Figure \ref{fig:maxima}). Furthermore, our theoretical analysis suggested that the residual error is driven by the error occurring in the situation $\tilde{X}_{MAP}\neq\tilde{X}_{MPM}$ since for $\tilde{X}_{MAP}=\tilde{X}_{MPM}$ the approximation error vanishes with an increasing number of samples. The segmentation uncertainty is quantified based on the Shannon entropy contained in the voxel-wise marginal distributions estimated by Perturb-and-MPM. As a consequence, the error in uncertainty quantification can also be bounded. Moreover, we emphasize that the estimated uncertainty is directly linked to the probabilistic model used for generating the segmentation itself. The computational overhead of the method is modest (sampling one segmentation corresponds to one inference step). Perturb-and-MPM can essentially be applied to any dense CRF that relies on a mean field approximation, e.g. \citep{Kamnitsas2016}.
\subsection{Comparison to other methods}
Methods used for sampling image segmentations can be grouped in two categories: Methods that perform sampling over the complete image grid, and methods that perform sampling from a parametric representation of the region of interest. The first group of methods relies conventionally on MCMC-based sampling approaches. They exhibit several disadvantages such as burn-in phase and typically bad scalability, which leads to a slow computation time \citep{Iglesias2013}. The computation time can be improved by employing parallelization techniques \citep{Byrd2010}. Additionally, MCMC-based sampling approaches usually rely on a number of sampling hyperparameters (e.g. step sizes, etc.) which need to be fine-tuned. In contrast, our method does neither require a burn-in phase nor does it require the tuning of any hyperparameters. The second group of methods typically employs curves as parametric representation for segmentation boundaries \citep{Fan2007,Le2016}. The parametric representation enables a more efficient sampling than in conventional MCMC-based approaches. However, these approaches do not handle sampling of multi-label segmentations naturally. In contrast, our method enables efficient sampling of multi-label segmentations over the complete image grid (corresponding to the CRF).

\citet{Kohli2008} proposed a method based on min-marginals for the computation of labeling uncertainty in pairwise, grid-structured CRFs. \citet{Parisot2014} used this technique within a combined segmentation-registration framework in order to segment binary tumor masks of low-grade glioma. \citet{Alberts2016a} used it in context of longitudinal tumor volumetry of gliomas. The estimation of min-marginals relies on the computation of st-cuts, which in case of multi-label problems is only possible for a very limited class of energy functions \citep{Ishikawa2003}. More importantly, their estimation is not possible for non-convex priors such as e.g. the Potts prior commonly used in image segmentation.
\subsection{Experimental aspects}
We investigated the quality and potential radiological utility of the uncertainty estimation for the multi-label segmentation of the tumor core in glioblastoma patients. We chose this particular segmentation task since recent benchmarks \citep{Menze2015} highlighted that the segmentation of the tumor core appears to be the most challenging segmentation task in glioblastoma for current state-of-the-art algorithms. In addition, the individual compartments of the tumor core have shown to be associated with molecular characteristics \citep{Gutman2015,Naeini2013} and patient survival \citep{Pope2005,Rao2016,RiosVelazquez2015}, emphasizing the importance of their accurate definition via automatic segmentation. Quality was defined in terms of sensitivity and specificity of the uncertainty estimation with regards to segmentation errors. We found that uncertain regions correspond well to wrongly segmented regions. However, we observed that uncertain voxels are mostly situated at the border/periphery of such a region, whereas central areas show less or no uncertainty (cf. Figure \ref{fig:NumberOfSamples} and \ref{fig:visual-examples}). Intuitively, it does make sense that an algorithm is most uncertain in delineating the interface of different tissues. This behavior can also be seen in other perturbation-based approaches \citep{Alberts2016,Papandreou2014}. In general, the apparent uncertainty is dependent on the marginal distributions of the model. Thus, one can observe that the spatial extent of segmentation uncertainty does not change anymore after a number of samples (cf. Figure \ref{fig:NumberOfSamples}). If the model at hand has very peaky distributions (i.e. is overconfident), the  corresponding uncertainty map will be less informative. As a consequence, wrongly labeled image regions may not be reflected in the corresponding uncertainty map. This observation brings interesting aspects for future developments, where the interplay between Perturb-and-MPM, to assess uncertainty, and the design as well as training of the underlying segmentation model, is to be considered.
\subsection{Clinical aspects}
The clinical utility of Perturb-and-MPM was demonstrated for the estimation of the EOR and RTV in glioblastoma patients. The EOR and RTV are important prognostic factors and hence their estimation is essential for patient prognosis \citep{Chaichana2014,Grabowski2014,Stummer2008}. We assessed the ability of Perturb-and-MPM to indicate uncertain voxels to be excluded from the computation of EOR and RTV, respectively. This resulted in a consistent improvement in the estimation of both parameters with respect to ground truth data of four expert raters. Consequently, we would expect also an improvement in association of such automatically extracted and corrected imaging biomarkers with patient survival. Similarly, we expect that these methods can contribute for a better assessment of disease progression and response to therapy, where a precise longitudinal assessment of tumor volumes is very important for clinical decision making \citep{Bauer2014a,Meier2016}. 
\subsection{Conclusion}
Perturb-and-MPM is an easy-to-implement approach for performing approximate sampling from dense multi-label CRFs. It allows the generation of spatially-resolved uncertainty maps for image segmentations. This opens up opportunities to create time-effective human-machine interfaces necessary to monitor and correct results of automated segmentation. More importantly, it can have a substantial impact in clinical scenarios such as radiotherapy and neurosurgery, where an accurate delineation of the tumor is needed in order to precisely target the tumor while preserving as much healthy tissues as possible. In turn, uncertainty estimates of the segmentation can be used as data quality measures in high-throughput radiomics studies relying on these segmentation results to extract imaging biomarkers of diagnosis, assessment of prognosis, and prediction of therapy response.

\subsubsection*{Funding} This project has received funding from the European Union’s Seventh Framework Programme for research, technological development and demonstration under grant agreement N\textordmasculine 600841 and the Swiss National Science Foundation project 205321\_169607.


\bibliography{MICCAI2016MPM}

\end{document}